\documentclass[acmsmall,nonacm]{acmart}
\AtBeginDocument{%
  }

\setcopyright{acmlicensed}
\copyrightyear{2018}
\acmYear{2024}
\acmDOI{XXXXXXX.XXXXXXX}

\acmJournal{JACM}
\acmVolume{37}
\acmNumber{4}
\acmArticle{111}
\acmMonth{8}

\usepackage{subcaption}
\usepackage{multirow}
\usepackage{algorithm}
\usepackage{algorithmic}
\usepackage{graphicx}
\usepackage{booktabs}

\begin{document}

\title{Advancing CMA-ES with Learning-Based Cooperative Coevolution for Scalable Optimization}

\author{Hongshu Guo}
\affiliation{%
  \institution{South China University of Technology}
  \city{Guangzhou}
  \state{Guangdong}
  \country{China}
}

\author{Wenjie Qiu}
\affiliation{%
  \institution{South China University of Technology}
  \city{Guangzhou}
  \state{Guangdong}
  \country{China}
}

\author{Zeyuan Ma}
\affiliation{%
  \institution{South China University of Technology}
  \city{Guangzhou}
  \state{Guangdong}
  \country{China}
}

\author{Xinglin Zhang}
\affiliation{%
  \institution{South China University of Technology}
  \city{Guangzhou}
  \state{Guangdong}
  \country{China}
}

\author{Jun Zhang}
\affiliation{%
 \institution{Nankai University, China; Hanyang University}
   \country{South Korea}
}

\author{Yue-Jiao Gong*}
\affiliation{%
  \institution{South China University of Technology}
  \city{Guangzhou}
  \state{Guangdong}
  \country{China}
}
\email{*Corresponding-Author: gongyuejiao@gmail.com}

\renewcommand{\shortauthors}{Trovato et al.}

\begin{abstract}

Recent research in Cooperative Coevolution~(CC) have achieved promising progress in solving large-scale global optimization problems. However, existing CC paradigms have a primary limitation in that they require deep expertise for selecting or designing effective variable decomposition strategies. Inspired by advancements in Meta-Black-Box Optimization, this paper introduces LCC, a pioneering learning-based cooperative coevolution framework that dynamically schedules decomposition strategies during optimization processes. The decomposition strategy selector is parameterized through a neural network, which processes a meticulously crafted set of optimization status features to determine the optimal strategy for each optimization step. The network is trained via the Proximal Policy Optimization method in a reinforcement learning manner across a collection of representative problems, aiming to maximize the expected optimization performance. Extensive experimental results demonstrate that LCC not only offers certain advantages over state-of-the-art baselines in terms of optimization effectiveness and resource consumption, but it also exhibits promising transferability towards unseen problems. 


\end{abstract}

\ccsdesc[500]{Mathematics of computing~Bio-inspired optimization}
\ccsdesc[500]{Computing methodologies~Reinforcement learning}

\keywords{CMA-ES, cooperative co-evolution, reinforcement learning, large scale global optimization, meta-black-box optimization}


\maketitle
\setlength{\floatsep}{1pt plus 1pt minus 2pt}
\setlength{\textfloatsep}{1pt plus 1pt minus 2pt}
\setlength{\intextsep}{1pt plus 1pt minus 2pt}
\section{Introduction}\label{intro}
Black box optimization (BBO) is a class of optimization problems whose objective function is either unknown or too intricate to be mathematically formulated \cite{ma2024metabox}. Consequently, BBO requires interaction-based information acquisition without access to underlying mathematical expressions or gradients. Within the context of BBO, Large-Scale Global Optimization (LSGO), which involves thousands to tens of thousands of variables, has numerous real-world applications \cite{elsken2019neural,dranka2021review,bhattacharya2016evolutionary} to drive resource savings, cost control, and efficiency enhancement \cite{10.1145/3236009,omidvar2021review,liu2024large,zhang2024survey}. Many works have proposed LSGO variants of algorithms originally applied to lower-dimensional BBO problems, such as Sep-CMAES\cite{ros2008simple}, LM-MA-ES\cite{loshchilov2018large}, and so on\cite{akimoto2016projection,loshchilov2017lm,he2020mmes,li2017simple}, to tackle such problems. Besides, Persistent Evolution Strategies (PES) \cite{vicol2021unbiased}, presented in an outstanding paper at ICML-2021, combines ideas from gradient-based optimization with evolutionary strategies to improve optimization efficiency and accuracy. 
However, the ``curse of dimensionality'' represents a significant challenge for such problems: as the number of variables increases, the complexity of optimization grows exponentially, necessitating extensive iterations for exploration \cite{hammer1962adaptive}. 


\begin{figure}[t]
    \centering
    \includegraphics[width=0.99\textwidth]{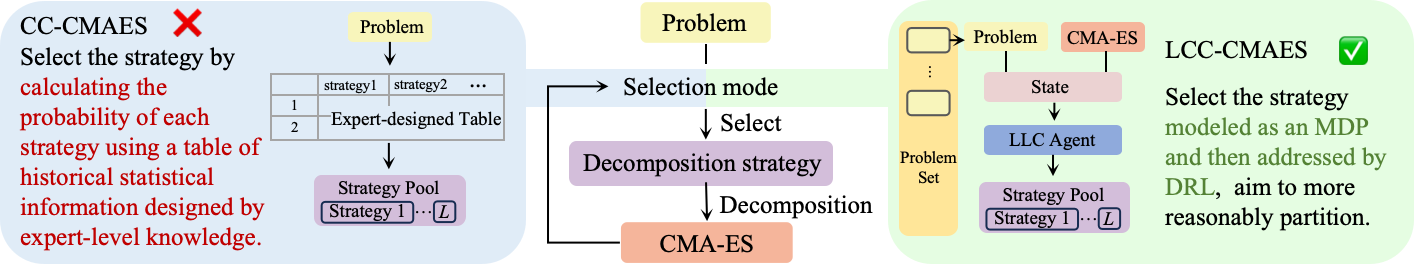}
    \caption{The core idea of LCC-CMAES.}
    \label{CCCMAESvsLCC}
\end{figure}

To address LSGO, inspired by the divide-and-conquer philosophy, a framework named Cooperative Co-evolution (CC) first divides the variables into several subgroups, then optimize these subgroups (considered as lower-dimensional BBO problems) using Evolutionary Algorithms (EAs), and finally integrates them into a comprehensive global optimization solution \cite{potter1994cooperative,jia2020contribution,chen2019cooperative,yang2008large}. In the CC framework, an important issue is placing non-separable variables within the same subgroup to accurately divide the problem dimensions, which is so called decomposition strategy \cite{van2004cooperative}. The researchers initially tried random decomposition and some decomposition strategies utilizing statistical data but did not obtain satisfactory results \cite{potter1994cooperative,van2004cooperative}. Later, they attempted to dynamically select strategies by calculating the probability of each using a table of historical statistical information, designed by expert-level knowledge, which yielded some positive effects (e.g., CC-CMAES \cite{liu2013scaling}). Furthermore, the researchers designed a series of decomposition strategies based on expert-level knowledge to more accurately identify variable interactions for precise decomposition, but this precise decomposition led to substantial additional function evaluations (FEs) costs \cite{sun2017recursive,omidvar2017dg2,tian2024composite}. According to the above,  a primary limitation in the current CC framework is the \textit{Expert-Level Knowledge Dependency}: these decomposition strategies are based on hand-crafted rules, heavily reliant on expert-level optimization knowledge and might not be generalizable towards unseen problems. Therefore, considering methods that do not require expert-level knowledge for decomposition could be a more suitable solution for tackling challenging real-world problems.


To alleviate the burdensome task of manual fine-tuning with expert-level knowledge, recent research has proposed the concept of Meta-Black-Box Optimization (MetaBBO) \cite{ma2024metabox,ma2024toward,li2024bridging,autosgnn,b2opt,li2024pretrained}. This paradigm has showcased the power of leveraging deep reinforcement learning (DRL) in a data-driven fashion at the meta-level to mitigate expert-level knowledge of low-level black-box optimizers. Numerous studies have shown that MetaBBO enables the black-box optimizers to achieve more effective optimization performance through enhanced parameter configuration \cite{xue2022multi,ma2024auto,sun2021learning}, algorithm/operator selection \cite{guo2024deep,liao2023two}, and update rule generation \cite{lange2023discovering,chen2024symbolic,chen2024symbol,yi2022automated}. Inspired by MetaBBO, we introduce \textbf{L}earning-Based \textbf{C}ooperative \textbf{C}oevolution (LCC), a pioneering framework that dynamically schedules decomposition strategies without expertise during optimization processes. The main contributions of this work are summarized as follows:

\begin{itemize}
    \item  
    LCC is designed to create an intelligent decision-making agent that autonomously selects effective decomposition strategies tailored to various problem environments and optimization states. We have formulated this process as a Markov Decision Process (MDP) and utilized DRL to construct the agent. This approach replaces traditional, expert-designed selection modes, marking a significant advance in automating and optimizing the decomposition strategy within the CC frameworks for large-scale BBO. 
    \item  Taking the Covariance Matrix Adaptation Evolution Strategy (CMA-ES) as the underlying optimizer, we develop the LCC-CMAES algorithm. Figure \ref{CCCMAESvsLCC} shows the core idea of LCC-CMAES. We have designed a set of straightforward yet representative statistical features to capture essential grouping information and reflect the optimization state. Based on the state, LCC selects an appropriate decomposition strategy from a strategy pool of random decomposition (RD), Min-Variance decomposition (MiVD) and Max-Variance decomposition (MaVD) - enhancing the efficacy of CMA-ES. 
    \item We conducted detailed comparisons with various leading LSGO algorithms in more challenging settings to illustrate the limitations in practical problems. The experimental results demonstrate that LCC-CMAES excels not only in terms of resource consumption but also in optimization results compared to other algorithms. Additionally, LCC-CMAES exhibits transferability, showing outstanding performance on other unseen problem sets after training. 
\end{itemize}

The remainder of this paper is organized as follows: Section \ref{Related Works} discusses related work. Section \ref{Preliminary} provides the preliminary knowledge necessary for understanding CC-CMAES and MDP. Section \ref{Methodology} describes the overall architecture of LCC, as well as the specific design of its MDP and network. Section \ref{Experiments} presents the experimental results and provides a detailed analysis. Finally, Section \ref{Conclusion and Futrue work} concludes the paper and outlines future work.

\section{Related Works}\label{Related Works}
As mentioned earlier, our LCC is inspired by MetaBBO and operates within the CC framework. Therefore, in this section, we will review MetaBBO and several important problem decomposition strategies under the CC framework.

\subsection{MetaBBO}
To alleviate the burdensome task of manual fine-tuning, the concept of MetaBBO has been proposed by recent research \cite{ma2024metabox,yang4956956meta,ma2024toward,metade,shao2025deep,faldor2025discovering}. MetaBBO aims to refine black-box optimizers by identifying optimal configurations or parameters through an automatic decision process without requiring expertise, thereby boosting overall performance across various problem instances within a given problem domain. MetaBBO-RL is one of approaches of MetaBBO \cite{sharma2019deep,tan2021differential,guo2025configx,ma2025accurate,ma2025surrogate}, which models the optimizer fine-tuning as a MDP and learns an RL agent to automatically make decisions without expertise. The meta-objective of MetaBBO-RL is to learn a policy (RL agent) $\Pi^{*}$  that maximizes the expectation of the accumulated meta-performance improvement $r_{t}$ (also called reward) over the problem set distribution $ \xi$, $\mathbb{E}_{\upsilon \sim \xi, \Pi^{*}}\left[\sum_{t=0}^{T} r_{t}\right]$ , where  $T$ denotes the all times of making decisions and $\upsilon$ is the problem of problem set $\Upsilon$. Specifically, in the aspect of operator selection, MetaBBO-RL automates the tuning process, significantly reducing the time and expertise needed to customize algorithms for specific unseen problems, while also potentially enhancing overall optimization performance \cite{xu2020reinforcement,wu2022employing,yin2021rlepso,guo2025reinforcement}. This has been confirmed in numerous research studies: RL-DAS \cite{guo2024deep}, based on MetaBBO-RL, selects operators for Differential Evolution algorithms, leveraging their complementary strengths to enhance optimization performance and demonstrating favorable generalization across different problem classes; RLDMDE \cite{yang2024dynamic} employs RL so that each subpopulation can adaptively select a mutation strategy based on the current environmental state (population diversity), thereby boosting the self-adaptation of subpopulations; similarly, RLEMMO \cite{lian2024rlemmo}, the first generalizable MetaBBO-RL framework for solving multimodal optimization problems (MMOP), selects operators for search strategies, directly addresses unseen problems, and achieves competitive optimization performance in both quality and diversity against several strong MMOP solvers.

\subsection{CC and the Problem Decomposition Strategies} \label{CC and the Problem Decomposition Strategie}
Inspired by the ``divide and conquer'' philosophy, CC is a framework to solve LSGO by the decomposition-based approach\cite{omidvar2021review2,omidvar2021review}. It first divides the variables into several subgroups, then optimize these subgroups using EAs, and finally integrates them into a global optimization solution.

CCGA \cite{potter1994cooperative} is the first strategy to use CC for problem decomposition, splitting an $D$-dimensional problem into $D$ one-dimensional problems, where $D$ is the dimensionality of the problem. However, both practical tests \cite{potter1994cooperative} and theoretical analyses \cite{van2004cooperative} have suggested that completely decomposing into one-dimensional problems poses a risk of introducing spurious minima. To mitigate this issue, strategies such as $k$-$s$ dimensional decomposition and bipartite decomposition have been proposed \cite{van2004cooperative,shi2005cooperative}, but these algorithms do not take into account the structure of the problem or interactions between variables, potentially placing interacting variables in different components, which adversely affects optimization performance \cite{omidvar2021review}. To achieve more precise decomposition, researchers have started from the definition of separability, defining various types of separability such as additive separability \cite{li2013benchmark}, multiplicative separability \cite{li2022dual}, and composite separability \cite{tian2024composite}, and have developed a range of variable interaction identification algorithms, such as DG2 \cite{omidvar2017dg2}, RDG \cite{sun2017recursive}, ERDG \cite{yang2020efficient}, MDG \cite{ma2022merged}, GDG \cite{mei2016competitive} and CSG \cite{tian2024composite}. 
In addition, researchers have further studied how to accurately decompose overlapping variables, such as DOV \cite{meselhi2022decomposition}, OCC \cite{komarnicki2024overlapping}, and OEDG \cite{tian2024enhanced}.
However, the cost of improving accuracy in this way includes a large number of expert-designed separability methods and additional FEs. Strategies based on probabilistic and statistical methods do not have these issues. They perform multiple rounds of grouping optimization before forming the final optimization result to capture problem structure and variable interactions \cite{yang2008large}. Relying on expertise, many algorithms were proposed \cite{tiwari2001interaction,roy2002generalised,tiwari2002variable,sobol1993sensitivity}, such as the the Delta method \cite{omidvar2010cooperative} based on theory that the improvement intervals for inseparable variables are relatively smaller than those for separable variables \cite{salomon1996re}, the Fitness Difference Minimization (FDM) method exemplified by DIMA \cite{sayed2012dependency} and CC-CMAES \cite{liu2013scaling} based on covariance matrices and expert-designed selection mode. Besides, contribution-based decomposition methods \cite{yang2023contribution}, such as CCFR \cite{yang2016efficient}, DCC \cite{zhang2019dynamic}, and CBCCO \cite{jia2020contribution}, represent another novel strategy. Although these two strategies do not have the additional FEs, they rely on expertise, so in different scenarios, they may fail to meet the requirements for reasonable decomposition\cite{omidvar2021review,qiu2025novel}. Therefore, considering methods that do not require expert-level knowledge for decomposition might be a more suitable solution for more challenging real-world problems.

\section{Preliminaries}\label{Preliminary}





\subsection{CC-CMAES}\label{cc-cmaes}

Covariance Matrix Adaptation Evolution Strategy (CMA-ES) \cite{hansen2016cma} is a representative EA that operates by repeatedly sampling offspring according to a distribution and updating the distribution with the performance of the sampled offspring until a stopping criterion is met (e.g., reaching the total number of generations $TG$).

\begin{equation}
    x_{(g+1)}^{(k)} \sim N\left(\omega_{(g)},\sigma_{(g)}^{2} \cdot C_{(g)}\right), \quad k=1,2, \cdots, \lambda
    \label{cmaes}
\end{equation}

Equation (\ref{cmaes}) shows the sampling process in a population $P$ with offspring size $\lambda$ at generation $g$. $\omega_{(g)} \in \mathbb{R}^D$, $C_{(g)} \in \mathbb{R}^{D\times D}$, and $\sigma_{(g)} \in \mathbb{R}$ are the Gaussian mean, covariance matrix, and global step size, respectively, at generation $g$. CC-CMAES \cite{liu2013scaling} uses the CC framework with CMA-ES, featuring three decomposition strategies: Min-Variance Decomposition (MiVD), Random Decomposition (RD), and Max-Variance Decomposition (MaVD), ranging from exploitative to exploratory. It dynamically selects one strategy to optimize subgroups with CMA-ES for a fixed number of generations until termination criteria are met. MiVD, RD, and MaVD decompose the space based on the rank of the diagonal of the covariance matrix. MiVD sequentially selects $D/m$ variables following the rank order to minimize the diversity among their variances. In contrast, MaVD selects one variable, then skips $D/m$ variables to select the next variable each time, which maximizes diversity. RD randomly selects $D/m$ variables within each subspace. The subspace covariance matrix \( C_{sub_i} \in \mathbb{R}^{(D/m) \times (D/m)}\) and mean \( \omega_{sub_i} \in \mathbb{R}^{D/m} \) are extracted from the global covariance matrix \( C \) and mean \( \omega \) as \( C_{sub_i} = C[subdims_i,subdims_i], \omega_{sub_i} = \omega[subdims_i] \), where \( subdims_i \in [1,D]^{D/m}\) represents the dimension index set of subgroup \( i \in \left[1, \dots, m \right] \). $C$ and $\omega$ are updated using $C_{sub_i}$ and $\omega_{sub_i}$ is its inverse process.


\subsection{Markov Decision Process}\label{Markov Decision Process And Proximal Policy Optimization}

A Markov Decision Process (MDP) is commonly characterized as $\mathcal{M}:=<\mathcal{S}, \mathcal{A}, \mathcal{T}, R>$. At each time step $t$, given the current environment state $s_{t} \in \mathcal{S}$, an action $a_{t} \in \mathcal{A}$ is performed according to a policy $\Pi:\mathcal{S} \rightarrow \mathcal{A}$. Then the environment reaches at the next state $s_{t+1}$ according to the transition dynamics $\mathcal{T}\left(s_{t+1} \mid s_{t}, a_{t}\right)$. The reward function $R: \mathcal{S} \times \mathcal{A} \rightarrow \mathbb{R}$ indicates the feedback $r_t$ from the environment. Given a finite horizon~(suppose $T$ steps) of interactions with the environment, a sampled trajectory is defined as $\tau:=\left(s_{0}, a_{0}, s_{1}, \cdots, s_{T}\right)$. Then an MDP is solved by finding an optimal policy $\pi^*$ that maximizes the expected accumulated rewards over all possible trajectories:
\begin{equation}
    \pi^{*}=\underset{\pi \in \Pi}{\arg \max } \mathbb{E}_{\tau \sim \pi(\tau)}[\sum_{t=0}^{T} \gamma^{t-1}r_t]
    \label{MDP}
\end{equation}
where $\pi(\tau)$ denotes the sampling probability of $\tau$ and $\gamma$ is a pre-defined discount factor. In the context of DRL~\cite{drl}, the policy $\pi$ is parameterized with a neural network $\pi_\theta$, which makes the gradient based learning methods~(e.g., PPO~\cite{ppo}) available for searching the optimal policy. 
\section{Methodology}\label{Methodology}
\subsection{LCC Overview}

\begin{figure}[t]
    \centering
    \includegraphics[width=0.8\textwidth]{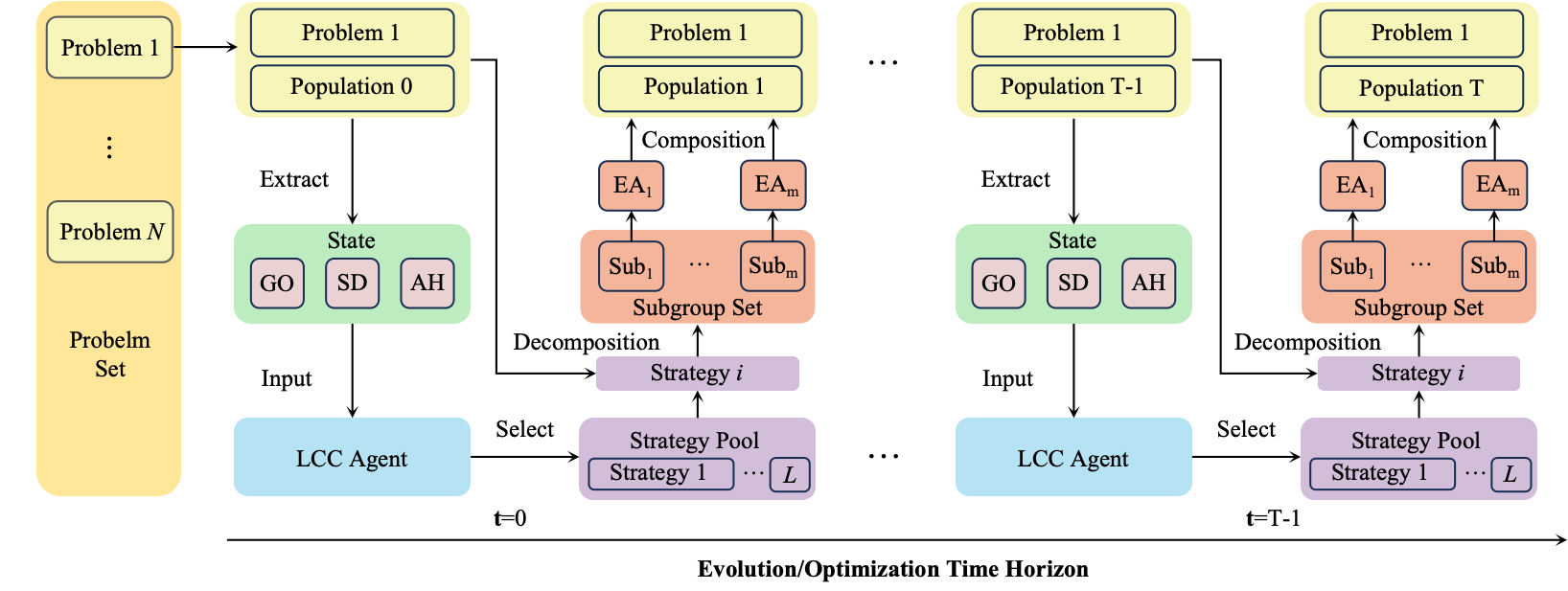}
    \caption{The overall structure of LCC.}
    \label{fig:The overall structure of LCC}
\end{figure}
\setlength{\floatsep}{1pt plus 1pt minus 2pt}
\setlength{\textfloatsep}{1pt plus 1pt minus 2pt}
\setlength{\intextsep}{1pt plus 1pt minus 2pt}

LCC primarily consists of three main components: a problem set $\Upsilon$ with $N$ problems, a CC decomposition strategy pool $\Lambda$, and an underlying EA optimizer (e.g., CMA-ES). $\Lambda$ includes a variety of strategies chosen from existing decomposition strategies. The detailed architecture, illustrated in Figure \ref{fig:The overall structure of LCC}, can be conceptualized as an MDP. In an MDP, as introduced in Section \ref{MDP}, multiple elements are fundamental, such as state $s_{t}\in \mathcal{S}$, action $a_{t}\in \mathcal{A}$, and reward $R: \mathcal{S} \times \mathcal{A} \rightarrow \mathbb{R}$. The DRL agent targets at the optimal policy $\pi^{*}$ that select an appropriate decomposition strategy in $\Lambda$ to maximize the expected accumulated reward over all the problems $\upsilon \in \Upsilon $ as $\pi^{*}=\underset{\pi \in \Pi}{\arg \max } \frac{1}{N}\sum_{k=1}^{N}\sum_{t=0}^{T} \gamma^{t-1} R\left(s_{t}, a_{t} | \upsilon\right)$. First, a problem $\upsilon$ is selected from the problem set $\Upsilon$. For this problem, we analyze Global Optimization (GO) information, Subgroup Decomposition (SD) information, and Action History (AH) information using Exploratory Landscape Analysis (ELA) \cite{mersmann2011exploratory} and Fitness Landscape Analysis (FLA) \cite{pitzer2012comprehensive}. This analysis is used to design the state to ensure it contains sufficient information to select an appropriate decomposition strategy. Based on the state $s_t$, LCC selects a decomposition strategy from the CC decomposition strategy pool $\Lambda$ to decompose the problem and subsequently optimize each subgroup using the underlying EA optimizer. A corresponding reward is designed to reflect improvements in the MDP $\mathcal{M}$. Finally, the optimized subgroups are combined to form a new global population, completing an epoch. With CMA-ES as the underlying optimizer, we instantiate LCC, naming it the LCC-CMAES algorithm. Using LCC-CMAES as a concrete example, we will describe the specific design of the MDP and the network.

\subsection{MDP Formulation}

\subsubsection{State}\label{sec:state}

The state space of LCC-CMAES encompasses the Global Optimization (GO) information $s_{\text{GO}}\in \mathbb{R}^{12}$, Subgroup  Decomposition (SD) information $s_{\text{SD}}\in \mathbb{R}^{4 \times m}$, and Action History (AH) information $s_{\text{AH}}\in \mathbb{R}^{2 \times L}$. Details are shown in the Table \ref{state}.

\begin{table}[t]
  \centering
  \scriptsize
  \caption{State features.}
  \renewcommand{\arraystretch}{1.5} 
  \resizebox{0.9\textwidth}{!}{%
  \begin{tabular}{c c c l}
    \toprule
    Feature & Feature Index & Calculation Formula & Explain \\
    \midrule
    \multirow{15}{*}{$s_{\text{GO}}$} 
    & 1 & $\text{Max}(\frac{\omega_t}{radius})$ & \multirow{3}{*}{\begin{tabular}[l]{@{}l@{}}
    $\text{Max}(\cdot),\text{Min}(\cdot)$ extracts the maximum and minimum element in the vector. $\text{Mean}(\cdot)$ extracts the mean \\of the vector elements. The $\omega_{t}$ is the global mean at step $t$. The $radius$ is the search radius of the problem,\\ which is half of the difference between the upper and lower bounds. To reflect the state of population \\ optimization and status of CMA-ES
    \end{tabular}} \\
    \cmidrule{2-3}
    & 2 & $\text{Mean}(\frac{\omega_t}{radius})$ &  \\
    \cmidrule{2-3}
    & 3 & $\text{Min}(\frac{\omega_t}{radius})$ &   
\\
    \cmidrule{2-4}
    & 4 &$\text{Max}(\text{Corrcoef}(C_{t}))$ & \multirow{3}{*}{\begin{tabular}[l]{@{}l@{}}
    $\text{Corrcoef}(\cdot)$ transforms a covariance matrix into a correlation coefficient matrix. $C_{t}$ is the global covariance \\ matrix at step $t$.  To reflect the correlations between  variables and status of CMA-ES
    \end{tabular}} \\
    \cmidrule{2-3}
    & 5 & $\text{Mean}(\text{Corrcoef}(C_{t}))$ & \\
    \cmidrule{2-3}
    & 6 & $\text{Min}(\text{Corrcoef}(C_{t}))$ &  \\
    \cmidrule{2-4}
    & 7 & $\frac{\sigma_{t}}{radius}$ & The $\sigma_{t}$ is the global step size at step $t$. To reflect the current global exploration and exploitation conditions\\
    \cmidrule{2-4}
    & 8 & $\text{Max}(\frac{gbest_{t}}{radius})$ & \multirow{3}{*}{\begin{tabular}[l]{@{}l@{}}
    The $gbest_{t}$ is the global best point at step $t$. To reflect the current
position of the optimization state within the \\ problem domain.
    \end{tabular}} \\
    \cmidrule{2-3}
    & 9 & $\text{Mean}(\frac{gbest_{t}}{radius})$ &  \\
    \cmidrule{2-3}
    & 10 & $\text{Min}(\frac{gbest_{t}}{radius})$ &  \\
    \cmidrule{2-4}
    & 11 & $\frac{f^*_{t}}{f^*_{t-1}}$ & The $f^*_{t}$ is the global best fitness at step $t$. To reflect the incremental optimization effects of each step.\\
    \cmidrule{2-4}
    & \multirow{2}[0]{*}{12} & \multirow{2}[0]{*}{$\frac{FEs}{MaxFEs}$} & $FEs$ is the number of remaining function evaluations, and $MaxFEs$ is the maximum number of function \\ &   &   & evaluations. To keep the agent informed about computational budget consumption\\
    \midrule
    \multirow{10}{*}{$s_{\text{SD}}$} 
    & 13-22 & $\text{Mean}(\text{Corrcoef}(C_{sub_i}))$ & $C_{sub_i}$ is the covariance matrix of subgroup $i$. To reflect the  correlation
between variables within the subgroup.\\
    \cmidrule{2-4}
    & \multirow{3}[0]{*}{23-32} & \multirow{3}[0]{*}{$\text{Mean}(\frac{\Delta subpop_i}{\lambda\times radius})$} &  The $\Delta subpop_i$ is the sum of the vector set consisting of the difference between the last generation and the first \\
      &   &   &  generation of each element of the subgroup. Inspired by the Delta method mentioned in Section \ref{CC and the Problem Decomposition Strategie},  \\&   &   & it aim to reveal interactions between variables. \\
    \cmidrule{2-4}
    & \multirow{3}[0]{*}{33-42} & \multirow{3}[0]{*}{$\text{Mean}(\frac{\text{Var}(subpop_i)}{radius^2})$} & $\text{Var}(\cdot)$ is calculated for the elements at each position within the vector set and $subpop_i$ is a vector set of all \\
      &   &   &  the generations of the subgroup. To reflect the volatility of the population 
across different  dimensions  within \\ &   &   &that subgroup and the exploration and exploitation of each dimension. \\
    \cmidrule{2-4}
    & \multirow{2}[0]{*}{43-52} & \multirow{2}[0]{*}{$\frac{d_{max_i}}{diameter}$} & The $d_{max_i}$ is the maximum distance in the subgroup $i$'s population. $diameter$ is the diameter of the search space.  \\
      &   &   &  To describe the convergence state of
the population within subgroup $i$. \\
    \midrule
    \multirow{7}{*}{$s_{\text{AH}}$} 
    & \multirow{3}[0]{*}{53-55} & \multirow{3}[0]{*}{$\frac{\sum (\Delta r_j)}{num_j}$} & $\sum(\cdot) $ is the sum of all values $\cdot$ for action $j$ and $j$ is the index of action, j = 1,2,3. The $\Delta r_j$ refers to the difference between \\
      &   &   & the reward obtained at step $t$   for action $j$ and the reward obtained at step $t-1$. The $num_j$ is the number of times \\
      &   &   &  action $j$ has been selected. To reflect the algorithm $j$'s contribution to optimization.\\
    \cmidrule{2-4}
    & \multirow{2}[0]{*}{56-58} & \multirow{2}[0]{*}{$\frac{\sum( \Delta gbest^{(j)})}{2\times radius \times num_j}$} & The $\Delta gbest^{(j)}$ refers to the Euclidean norm of the difference between $gbest_t$ obtained at step $t$ for action $j$ \\
      &   &   & and $gbest_{t-1}$ obtained at step $t-1$. To reflect the algorithm $j$’s effectiveness in optimization.\\
    \bottomrule
  \end{tabular}%
  }
  \label{state}%
\end{table}%

For $s_{\text{GO}}$, reflects the CMA-ES state and global optimization state, revealing the complexity and difficulty of the optimization problem as well as the relationships between various dimensions. For $s_{\text{SD}}$, we have designed four types of features based on probabilistic and statistical methods within the CC framework to reflect the variable grouping status within a subgroup, which provides detailed insights into the dynamics of variable relationships and optimization progress in the subgroups. For $s_{\text{AH}}$, given that LCC includes a CC decomposition strategy pool $\Lambda$, we derive $s_{\text{AH}}$ to provide the RL agent with additional contextual knowledge about the optimization capabilities of the candidate strategies. 

Finally, the complete state in the MDP of LCC-CMAES is the integration of $s_{\text{GO}}$, $s_{\text{SD}}$ and $s_{\text{AH}}$. 
\begin{equation}
    \text { state }:=\left\{s_{\text{GO}} \in \mathbb{R}^{12}, s_{\text{SD}} \in \mathbb{R}^{4 \times m},s_{\text{AH}} \in \mathbb{R}^{2 \times L}\right\}
\end{equation}
Here, \( m \) represents the number of subgroups (where $m$ is 10 in LCC-CMAES), and \( L \)  denotes the number of CC decomposition strategies in the pool (where $L$ is 3 in LCC-CMAES). 


\subsubsection{Action}\label{sec:action}
We designed a strategy pool $\Lambda$ in advance, containing various decomposition strategies for selection. LCC selects a CC decomposition strategy from $\Lambda$ based on the state to achieve dynamic decomposition. For the purpose of balancing exploration and exploitation, LCC-CMAES utilize three types of decomposition strategies \cite{liu2013scaling}: MiVD, RD, and MaVD as introduced in Section~\ref{cc-cmaes}. This operation is represented as an integer, which indicates the index of the chosen strategy within the strategy pool of $L$ candidate strategies, denoted as \( a \in [1, L] \). Next, based on the selected strategy, the problem is divided into smaller-dimensional subproblems and then optimized using CMA-ES. The optimization results of each subproblem are subsequently combined into a global optimization result.
\subsubsection{Reward}\label{sec:reward}
To guide the agent towards achieving a lower cost, the reward function should consider the absolute reduction in cost at each time step $t$:

\begin{equation}
    r_{t}=\frac{f _{t-1}^{*}-f _{t}^{*}}{f_{0}^{*} - f^*}
    \label{reward}
\end{equation}

where $f _{t-1}^{*}$ and $f _{t}^{*}$ are the global best fitness in the $t$-$1$ step and the $t$ step. $f^*$ is the optimal fitness of the problem, $f_{0}^{*}$ is the global best fitness in the initial population, which serves as a normalization factor. This measures the performance improvement brought in the step $t$ optimization.
\subsection{Network Design}\label{Network}
\begin{figure}[t]
    \centering
    \includegraphics[width=0.8\textwidth]{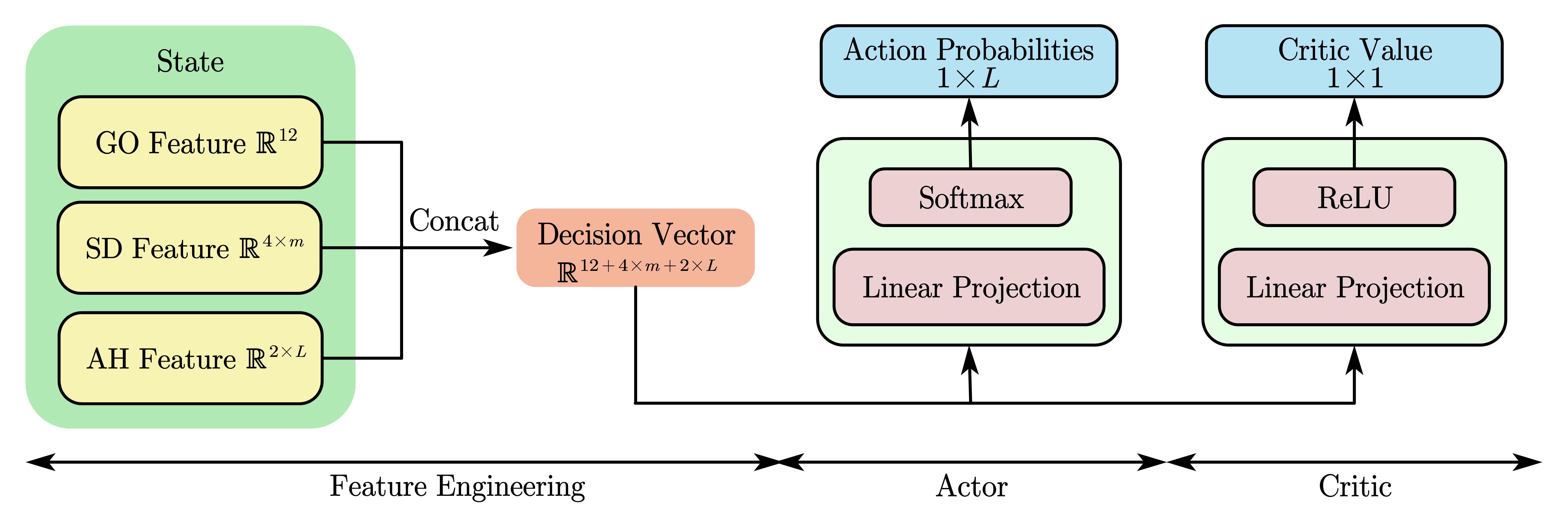}
    \caption{The Neural Network workflow for $\pi_{\theta}$ (Actor) and $ v_{\phi}$(Critic).}
    \label{fig:NN}
\end{figure}

As shown in Figure \ref{fig:NN}, the network consists of three modules: Feature Processing, Actor, and Critic. $s_{\text{GO}}$, $s_{\text{SD}}$, and $s_{\text{AH}}$ are first fused to form a state representation vector $DV$. Based on this representation, the Actor outputs the probability distribution of candidate strategies, while the Critic estimates the return value.

The Actor decides the probability for selecting a strategy from the CC decomposition strategies pool. 
As mentioned in Section~\ref{sec:state}, LCC-CMAES has $m=10$ subgroups and $L=3$ actions. We first concatenate $s_\text{GO} \in \mathbb{R}^{12}$, $s_\text{SD} \in \mathbb{R}^{4\times m}$ and $s_\text{AH} \in \mathbb{R}^{2\times L}$ to generate the Decision Vector $DV \in \mathbb{R}^{58}$ as $DV = s_{\text{GO} }\oplus  s_{\text{SD}} \oplus s_{\text{AH}}$. Then we map $DV$ to a three-layer Multi-Layer Perceptron (MLP) network with the structure ($58 \times 64 \times 64 \times L$), cooperating with a ReLU~\cite{relu} activation after each hidden layer. Following the Softmax operation, the Actor outputs a probability distribution over the  strategy pool $\Lambda$, which is then used to sample the strategy.

The Critic also takes $DV$ as input and uses the same MLP structure as Actor, where the output dimension is set to be $1$ for critic value prediction. However, their MLP parameters are not shared, and the training is conducted independently.


\subsection{Workflow}\label{Work and Alpha Code}

LCC-CMAES's workflow begins with selecting a problem $\upsilon$ from the problem set $\Upsilon$, initializing the global dimension $D$, global covariance matrix $C_0 = I$, global population $P_0$, global step size $\sigma_0 = radius$ and global mean vector $\omega_0$. Then training for $\upsilon$ starts and terminates when $MaxFEs$ is exhausted or the global best fitness $gbest_t$ is lower than the termination error. After initialization, an MDP starts. At step $t$, state $s_t$ can be calculated by following Table \ref{state}. Based on $s_t$, the Actor policy \( \pi_\theta \) with parameters $\theta$ takes the Decision Vector $s_t$ as input and outputs the probability distribution of candidate strategies $\pi(a_t|s_t)$, while the the critic network $v_{\phi}$ with parameters $\phi$ predicts the expected return values (accumulated rewards) of $s_t$.  Once the strategy is determined, the problem is decomposed into subgroups, marking the end of the CC problem decomposition layer and transitioning into the subgroup optimization layer. Each subgroup is optimized using CMA-ES until \( SubMaxFEs \) is reached, with \( \sigma_t \) updated by the offspring in each subgroup. Once all subgroup optimization completed, \( \sigma_t \) is updated to \( \sigma_{t+1} \), \( C_t \), \( \omega_t \), and \( P_t \) are updated to \( C_{t+1} \), \( \omega_{t+1} \), and \( P_{t+1} \), using on the \( C_{sub_i} \) and \( \omega_{sub_i} \) obtained from each subgroup, as mentioned in Section \ref{cc-cmaes}. At this point, state $s_{t+1}$ can be calculated by following Table \ref{state}. Then the reward $r_t$ is observed. 
The trajectories of states $s_t$, actions $a_t$, and rewards $r_t$ are recorded and then used by the PPO method to train the policy net $\pi_\theta$ and the critic net $v_{\phi}$ for $K$ times after the completion of optimization. PPO is trained in an actor-critic manner. It proposes a novel objective with clipped probability ratios, which
forms a first-order estimate (i.e., lower bound) of the policy’s performance. Its objective function at the $k$-th learning iteration ($k \in [1, K]$) is defined as: $L_{\pi}(\theta^{(k)}) := \mathbb{E} \left[ \min \left( \eta(\theta^{(k)}) \hat{A}, \operatorname{clip}(\eta(\theta^{(k)}), 1-\epsilon, 1+\epsilon) \hat{A} \right) \right]$ where $\eta^{(k)} := \frac{\pi_{\theta^{(k)}}(a_t|s_t)}{\pi_{\theta^{(0)}}(a_t|s_t)}$ is the ratio of the probabilities under the current policy and the old policy before the $K$-step learning process, performing the importance sampling. $\hat{A}$ is the estimated advantage calculated as the difference between the target return $G$ and the estimated return $\hat{G}$. Using $L_{\pi}(\theta)$, the gradients are back-propagated through the network to update the parameters and achieve the training effect. The critic network $v_{\phi}$ takes the Decision Vector as input and outputs a critic value prediction to estimate the return value $\hat{G}$. The loss function of the critic network $v_{\phi}$ is: $L_{v}(\phi) := \operatorname{MSE}(G, \hat{G})$.

\section{Experiments}\label{Experiments}

\subsection{Experimental Setup}
\subsubsection{Comparison Algorithms for LCC-CMAES }
For a comprehensive comparisons, we selected CC-CMAES \cite{liu2013scaling}, CSG \cite{tian2024composite}, ERDG \cite{yang2020efficient}, MDG \cite{chen2022decomposition}, and FII \cite{ge2015cooperative} as the comparison algorithms under CC framework. We then selected LSGO baselines without CC: CMA-ES \cite{hansen2016cma}, Sep-CMAES \cite{ros2008simple}, LM-CMA \cite{loshchilov2017lm}, LM-MA-ES \cite{loshchilov2018large}. Besides, MetaBBO method MetaES \cite{lange2023discovering} and local search method L-BFGS \cite{byrd1995limited} were also chosen. Among the comparison algorithms under CC framework, CC-CMAES, having the same decomposition strategy pool, is introduced in Section \ref{cc-cmaes}. CSG, ERDG, MDG, FII are the algorithms that need addition FEs costs for decomposition: CSG is currently the most powerful multi-stage variable identification algorithm; ERDG is a more efficient variant of RDG3 \cite{sun2019decomposition} (the 2018 CEC LSGO champion); MDG is an algorithm that addresses overlapping problems; FII is a rapid identification algorithm that reduces the FEs consumed by decomposition. These algorithms under CC framework all use CMA-ES as the underly optimizer. Among other types of comparison algorithms, MetaES discovers evolutionary strategies via MetaBBO and serves as the global optimization comparison algorithm within MetaBBO; L-BFGS is an optimization algorithm from outside the evolutionary algorithms community, and it is widely used in practical applications, especially for LSGO.

\subsubsection{Benchmark and Hyperparameter Settings} \label{Benchmark and Hyperparameter Settings}
The CEC 2013 LSGO benchmark \cite{li2013benchmark} comprises a total of $N=15$ problems, which are divided into five types of functions: Fully-separable Functions (F1-F3), Partially Additively Separable Functions with a separable subcomponent (F4-F7), Partially Additively Separable Functions with no separable subcomponents (F8-F11), Overlapping Functions (F12-F14), and Non-separable Function (F15).  Additionally, we partitioned the CEC 2013 LSGO benchmark suite, as shown in Table \ref{comparison_algorithms}. An asterisk ``*'' marks the problems used for training, while the rest were used for testing. Except for the Non-separable Function, each category has training problems, and F1, F4, and F8 are variants of the Elliptic Function; F5 and F9 are variants of the Rastrigin Function. This allows for testing LCC-CMAES's generalizability on unseen functions within the same type, with the Non-separable Functions tested as the unseen functions and type. 


The hyperparameter settings in this paper are as follows: the
total number of generations ($TG$) is 50, the offspring size ($\lambda$) is 20, subgroup maximum function evaluations ($SubMaxFEs$) is 1E3, the number of subgroups ($m$) is 10, learning rate ($lr$) is 6E-4, number of epochs ($Epoch$) is 90, Mini PPO iterations ($K$) is 3, the problem's dimension ($D$) is 1E3 (F13 and F14 in CEC2013LSGO is 905), the number of action selecting ($ns$) is 20. It’s worth noting that,
to realistically address the practical problems and accommodate the extra decomposition costs required by
various decomposition strategies, we set $MaxFEs = TG \times \lambda \times m \times ns = 50 \times 20 \times 10 \times 20 = $ 2E5. The settings of the comparison algorithms remain the same as those in their original papers.

\subsection{Comparison Analysis}

\begin{table}[t]
  \centering
  \caption{Comparing LCC-CMAES with comparison algorithms on CEC 2013 LSGO.}
  \setlength{\tabcolsep}{4pt} 
  \renewcommand{\arraystretch}{0.8} 
  \resizebox{\textwidth}{!}{ 
    \begin{tabular}{r|cccccc|cccccc}
    \hline
    \multirow{2}{*}{problem} & \multicolumn{6}{|c|}{Algorithms under CC Framework} & \multicolumn{6}{c}{Algorithms without CC} \\  \cline{2-13}
     & LCC-CMAES & CC-CMAES & CSG & ERDG & MDG & FII & CMA-ES & Sep-CMAES  & LM-CMA & LM-MA-ES  & MetaES  & L-BFGS \\
    \hline
    \multicolumn{1}{c|}{*1} & 2.045E+07 & 3.244E+08(+) & 4.182E+08(+) & 4.227E+08(+) & 4.182E+08(+) & 4.227E+08(+) & 4.223E+08(+) & \textbf{8.591E+06}(-) &  2.313E+07(+) & 4.691E+07(+) & 2.420E+11(+) & 9.085E+09(+) \\
      & $\pm$2.781E+06 & $\pm$1.112e+08 & $\pm$3.882E+07 & $\pm$3.825E+07 & $\pm$3.884E+07 & $\pm$2.231E+07 & $\pm$3.827E+07 & $\pm$\textbf{1.777e+06} & $\pm$4.576E+06 & $\pm$2.973E+06 & $\pm$2.351E+10 & $\pm$1.396E+08 \\
    \multicolumn{1}{c|}{2} & 4.419E+03 & \textbf{2.159E+03}(-) & 2.636E+03(-) & 2.555E+03(-) & 2.636E+03(-) & 2.592E+03(-) & 5.108E+03(+) & 5.414E+03(+) & 1.949E+04(+) & 6.886E+03(+) & 4.789E+04(+) & 4.050E+04(+)\\
      & $\pm$2.022E+02 & $\pm$\textbf{4.125E+02} & $\pm$1.393E+02 & $\pm$1.294E+02 & $\pm$1.397E+02 & $\pm$9.076E+01 & $\pm$2.32E+02 & $\pm$4.213E+01 & $\pm$1.483E+03 & $\pm$2.674E+02  & $\pm$5.721E+03 & $\pm$3.281E+03\\
    \multicolumn{1}{c|}{3} & \textbf{2.007E+01} & 2.036E+01(+) & 2.162E+01(+) & \textbackslash{}(+) & 2.161E+01(+) & 2.161E+01(+) & 2.161E+01(+) & 2.11E+01(+) & 2.052E+01(+) & 2.172E+01(+) & 2.171E+01(+) & 2.165E+01(+)\\
      & $\pm$\textbf{4.182E-02} & $\pm$2.289E-02 & $\pm$4.061E-03 & \textbackslash{} & $\pm$7.218E-03 & $\pm$2.374E-01 & $\pm$4.197E-02  & $\pm$2.394E-02 & $\pm$3.692E-02 & $\pm$3.486E-02 & $\pm$7.976E-01 & $\pm$7.686E-02 \\
    \multicolumn{1}{c|}{*4} & 6.932E+10 & 1.902E+11(+) & 3.502E+10(-) & 4.561E+10(-) & 3.270E+10(-) & 2.603E+12(+) & 2.681E+12(+) & 1.632E+11(+) & \textbf{2.492E+10(-)} & 1.220E+11(+) & 2.440E+12(+) & 5.084E+12(+) \\
      & $\pm$9.792E+09 & $\pm$8.122E+10 & $\pm$3.645E+09 & $\pm$3.218E+09 & $\pm$6.397E+09 & $\pm$1.218E+12 & $\pm$7.639E+11 & $\pm$2.938E+10  & $\pm$\textbf{1.009E+09} & $\pm$8.678E+10  & $\pm$5.409E+11  & $\pm$5.728E+11\\
    \multicolumn{1}{c|}{*5} & 5.386E+06 & 9.455E+06(+) & \textbf{1.079E+06}(-) & 2.053E+06(-) & 1.146E+06(-) & 1.172E+06(-) & 4.205E+06($\approx$) & 2.274E+06(-) 
 & 9.370E+06(+) & 1.534E+06(-) & 4.976E+07(+) & 4.890E+07(+) \\
      & $\pm$1.982E+06 & $\pm$2.514E+06 & $\pm$\textbf{1.485E+05} & $\pm$2.363E+05 & $\pm$1.938E+05 & $\pm$1.845E+05 & $\pm$2.954E+05 & $\pm$4.255E+05 & $\pm$1.295E+06 & $\pm$1.126E+05 & $\pm$5.214E+06 & $\pm$2.386E+06 \\
    \multicolumn{1}{c|}{6} & 1.048E+06 & 1.056E+06($\approx$) & 1.066E+06(+) & \textbackslash{}(+) & 1.066E+06(+) & 1.065E+06(+) & 1.062E+06(+) & 1.078E+06(+)  & \textbf{1.038E+05($\approx$)} & 1.063E+06(+) & 1.000E+06(-) & 1.071E+06(+)\\
      & $\pm$2.147E+03 & $\pm$4.632E+03 & $\pm$7.156E+02 & \textbackslash{} & $\pm$5.673E+02 & $\pm$8.675E+02 & $\pm$1.287E+03 & $\pm$2.745E+03 & $\pm$\textbf{1.703E+03} & $\pm$6.845E+03 & $\pm$8.232E+03 & $\pm$3.219E+04\\
    \multicolumn{1}{c|}{7} & 7.306E+08 & 3.268E+09(+) & 2.140E+07(-) & 5.059E+07(-) & \textbf{7.295E+06}(-) & 4.442E+09(+) & 6.739E+08($\approx$) & 1.756E+09(+)  & 3.080E+08(-) & 2.871E+07(-) & 3.240E+14(+) & 1.210E+15(+)\\
      & $\pm$2.502E+08 & $\pm$2.404E+09 & $\pm$8.465E+06 & $\pm$8.356E+06 & $\pm$\textbf{4.113E+06} & $\pm$1.532E+09 & $\pm$2.035E+08 & $\pm$7.201E+08 & $\pm$2.736E+07 & $\pm$4.212E+06 & $\pm$7.248E+13 & $\pm$5.317E+10\\
    \multicolumn{1}{c|}{*8} & 2.299E+15 & 1.547E+16(+) & 3.488E+15($\approx$) & 1.543E+16(+) & 1.513E+15($\approx$) & 2.885E+15($\approx$) & 6.162E+16(+) & 3.184E+15(+) & \textbf{1.061E+13(-)} & 2.189E+15($\approx$) & 1.940E+16(+) & 1.460E+17(+)\\
      & $\pm$2.047E+15 & $\pm$7.053E+15 & $\pm$2.872E+15 & $\pm$6.205E+15 & $\pm$1.194E+15 & $\pm$1.413E+15 & $\pm$1.492E+16 & $\pm$7.612E+14 & $\pm$\textbf{4.268E+12} & $\pm$1.062E+15 & $\pm$1.382E+16 & $\pm$2.014E+15\\
    \multicolumn{1}{c|}{*9} & 5.818E+08 & 8.084E+08(+) & 6.199E+08(+) & 1.443E+09(+) & 5.104E+08($\approx$) & 6.007E+08($\approx$) & 6.812E+08(+) & 3.890E+08(-) & 6.380E+08(+) & \textbf{3.020E+08}(-)  & 3.270E+09(+) & 3.730E+09(+)\\
      & $\pm$1.195E+08 & $\pm$1.636E+08 & $\pm$2.793E+07 & $\pm$1.572E+08 & $\pm$2.175E+08 & $\pm$4.263E+07 & $\pm$1.751E+07 & $\pm$3.527E+07 & $\pm$2.165E+06 & $\pm$\textbf{3.214E+07} & $\pm$6.832E+07 & $\pm$5.833E+08\\
    \multicolumn{1}{c|}{10} & 9.423E+07 & 9.375E+07($\approx$) & 9.464E+07($\approx$) & 9.576E+07(+) & 9.538E+07(+) & 9.523E+07(+) & 9.464E+07($\approx$) & 9.447E+07($\approx$) & \textbf{9.062E+07}(-) & 9.829E+07(+) & 9.803E+07(+) & 9.696E+07(+)\\
      & $\pm$5.623E+05 & $\pm$6.224E+05 & $\pm$2.725E+05 & $\pm$1.653E+05 & $\pm$1.712E+05 & $\pm$1.332E+05  & $\pm$1.426E+05 & 1.012E+05 & $\pm$\textbf{1.191E+05} & $\pm$3.321E+05  & 3.115E+05 & $\pm$2.133E+05\\
    \multicolumn{1}{c|}{11} & 8.243E+09 & 2.297E+11(+) & 2.998E+17(+) & 3.609E+17(+) & 4.770E+17(+) & 6.327E+17(+) & 2.364E+10(+) & 1.850E+10(+) & \textbf{5.620E+08}(-) & 1.720E+09(-) & 6.630E+22(+) & 6.470E+16(+) \\
      & $\pm$6.352E+09 & $\pm$6.313E+10 & $\pm$1.245E+15 & $\pm$1.431E+16 & $\pm$1.586E+15 & $\pm$8.848E+14 & $\pm$4.967E+09 & $\pm$2.786E+08  & $\pm$\textbf{3.921E+07} & $\pm$2.179E+08  & $\pm$8.795E+21 & $\pm$2.545E+15 \\
    \multicolumn{1}{c|}{*12} & 2.135E+03 & 1.766E+05(+) & 2.497E+05(+) & 4.140E+06(+) & 1.321E+03(-) & 1.079E+03(-) & 1.103E+03(-) & \textbf{1.068E+03(-)} & 2.157E+03($\approx$) & 1.079E+03(-) & 6.860E+12(+) & 1.040E+08(+)\\
      & $\pm$4.072E+02 & $\pm$1.484E+04 & $\pm$6.373E+04 & $\pm$1.464E+06 & $\pm$4.063E+02 & $\pm$5.374E+01 & $\pm$8.365E+01 & $\pm$\textbf{1.476E+02} & $\pm$5.921E+01 & $\pm$1.343E+02 & $\pm$3.215E+12 & $\pm$2.354E+07\\
    \multicolumn{1}{c|}{*13} & 1.298E+10 & 2.730E+10(+) & 2.332E+11(+) & 8.073E+15(+) & 1.747E+10($\approx$) & 3.274E+12(+) & 7.857E+09(-) & 1.698E+10($\approx$) & 5.530E+09(-) & \textbf{8.330E+08(-) }& 2.100E+21(+) & 9.440E+16(+)\\
      & $\pm$2.643E+09 & $\pm$8.185E+09 & $\pm$4.099E+10 & $\pm$1.785E+16 & $\pm$5.493E+09 & $\pm$2.203E+12 & $\pm$1.964E+09 & $\pm$1.034E+10 & $\pm$1.353E+09 & $\pm$\textbf{2.573E+08}   & $\pm$7.595E+20 & $\pm$5.221E+15\\
    \multicolumn{1}{c|}{14} & 1.323E+11 & 3.761E+11(+) & 6.072E+11(+) & 3.599E+13(+) & 5.212E+21(+) & 2.953E+13(+) & \textbf{5.600E+08}(-) & 2.081E+10(-) & 1.09E+10(-) & 1.110E+10(-) & 8.990E+08(-) & 5.350E+22(+) \\
      & $\pm$3.895E+10 & $\pm$1.856E+11 & $\pm$1.253E+11 & $\pm$1.415E+13 & $\pm$7.284E+21 & $\pm$2.512E+13 & $\pm$\textbf{1.545E+10} & $\pm$1.461E+10 & $\pm$5.999E+09 & $\pm$1.496E+08 & $\pm$7.818E+21 & $\pm$2.527E+17 \\
    \multicolumn{1}{c|}{15} & \textbf{3.090E+07} & 3.306E+07($\approx$) & 1.508E+08(+) & \textbackslash{}(+) & 9.125E+07(+) & 9.612E+07(+) & 9.565E+07(+) & 4.67E+08(+) & 4.752E+07(+) & 3.760E+07(+) & 8.230E+15(+) & 2.350E+15(+) \\
      & $\pm$\textbf{5.235E+06} & $\pm$4.994E+06 & $\pm$1.593E+07 & \textbackslash{} & $\pm$7.689E+06 & $\pm$1.317E+07 & $\pm$1.628E+07 & $\pm$7.041E+06  & $\pm$2.128E+06 & $\pm$2.507E+06  & $\pm$6.947E+14 & $\pm$5.319E+14\\
    \hline
      & NA & 11/3/1 & 8/4/3 & 11/0/4 & 6/4/5 & 10/2/3 & 9/3/3 & 8/2/5 & 6/2/7 & 7/1/7 & 13/0/2 & 15/0/0\\
    \hline
    \end{tabular}%
    }
  \label{comparison_algorithms}
\end{table}%

\subsubsection{Comparison With Other Algorithms}\label{Comparison With Other Algorithm}


Table \ref{comparison_algorithms} presents the mean optimization results from 25 independent runs for each algorithm. The symbols ``+'', ``-'', and ``$\approx$'' denote the outcomes of the Wilcoxon rank-sum test at the 0.05 significance level, indicating whether the competing method performed better (+), worse (-), or showed no significant difference ($\approx$) compared to LCC-CMAES. The last column shows the test results for each algorithm, listing the number of times LCC-CMAES significantly outperformed competitors, instances with no significant difference, and cases where LCC-CMAES performed worse.


Based on the results from Table \ref{comparison_algorithms}, we can analyze the following outcomes:

\begin{itemize}
    \item \textbf{Superior Performance Within CC Frameworks}: LCC-CAMES demonstrates significant advantages compared to existing advanced algorithms within the CC framework. LCC-CMAES shows pronounced superiority on more challenging grouping problems (such as F12-F14 Overlapping Functions), highlighting its capability to handle more complex real-world problems.
    \item \textbf{Generalization Capability}: LCC-CMAES exhibits a degree of generalizability, thanks to the well-designed state that provides it with ample information. It shows certain generalization on similar problem types, such as F11, achieving commendable results when trained on F8-F9. Moreover, LCC-CMAES also demonstrated generalizability on completely unseen problem types (F15). This reveals LCC's ability to solve more complex real-world problems.
    \item \textbf{Improvement Over CC-CMAES}: LCC-CMAES also shows significant improvements compared to CC-CMAES, which employs the same decomposition strategy pool. This improvement is attributed to our effective design of the reward and state, which encourage more rational decomposition strategies and superior outcomes.
    \item \textbf{Comparison of Baselines Without CC}: LCC-CMAES surpasses CMA-ES, LM-MA-ES, MetaES, L-BFGS and shows competitive performance with LM-CMA which validates the effectiveness of LCC-CMAES.  
    
\end{itemize}

\subsubsection{Comparison on Extended Optimization Horizon}\label{exp:3e6}

To investigate the performance of algorithms under extended optimization horizons, we present the performance curves of the baselines on the 15 problems of CEC2013LSGO with 3E6 $MaxFEs$ in Figure \ref{CSG, CC-CMAES, ERDG, and LCC at Different FEs}. 
In most problems, LCC-CMAES performs better than CC-CMAES, demonstrating the universality of action selection effectiveness. In the early and middle stages of decomposition, LCC-CMAES outperforms other algorithms under the CC. However, after decomposition is complete, this advantage gradually diminishes and may even be surpassed. In most problems, except for L-BFGS and MetaES, which show poorer optimization results, algorithms without CC exhibit more prominent optimization performance. Algorithms under the CC perform better on separable problems (F4-F11), but struggle to show significant optimization effects (stepwise decline) on overlapping (F13-F14) and fully non-separable problems (F15). These problems are difficult to decompose, so all dimensions are often treated as a whole during optimization. This will also result in substantial resource consumption, as CMAES struggles with LSGO (e.g., the time cost of ERDG on F13 under 3E6 FEs reaches 60,000 seconds, whereas LCC-CMAES requires only 1,200 seconds). LCC-CMAES is not subject to such limitations, as it continues to perform decomposition even in overlapping and fully non-separable problems, hence significantly surpasses CC methods on these problems.


\begin{figure}[t]
    \centering
    \includegraphics[width=0.9\textwidth]{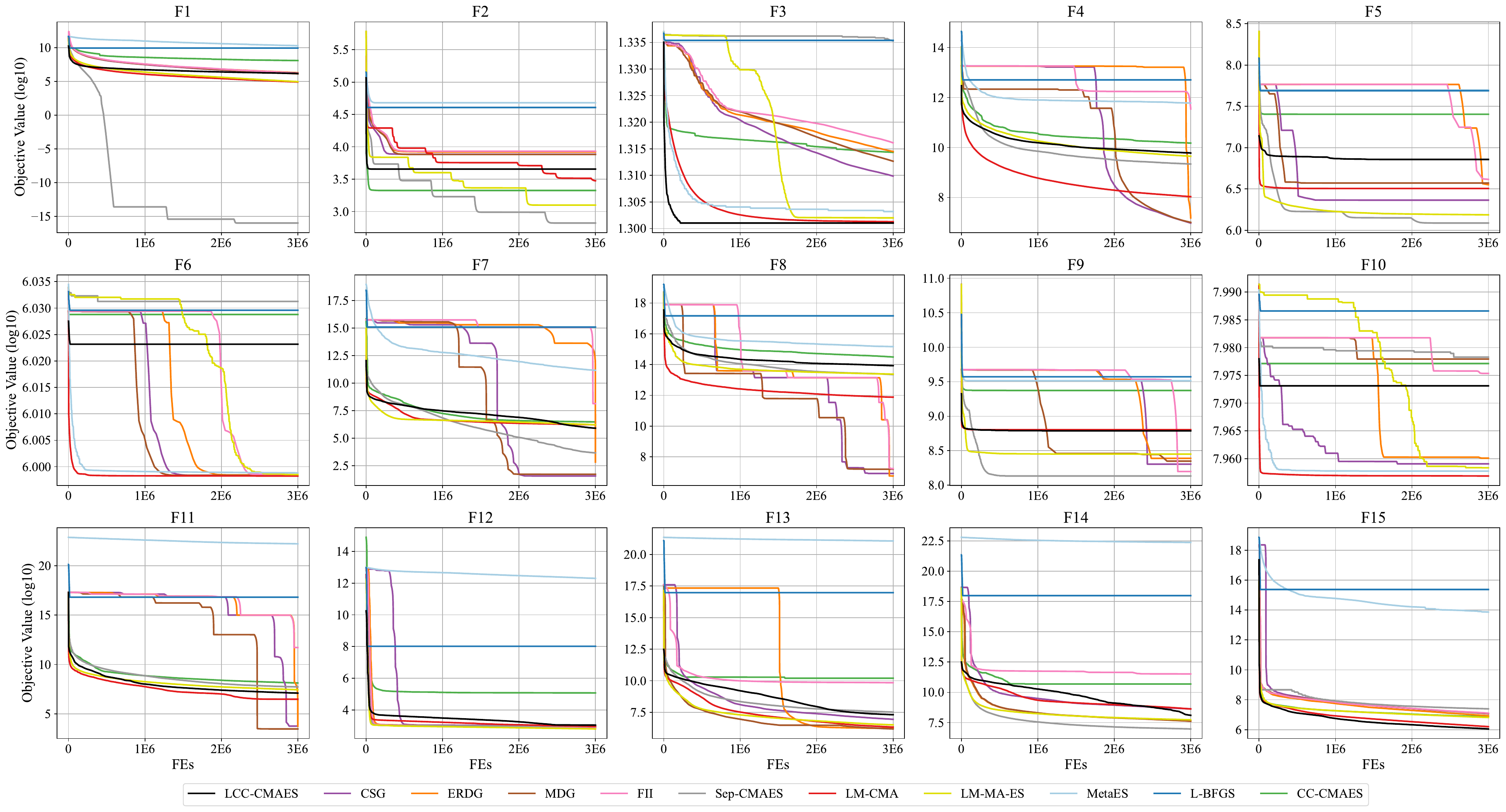}
    \vspace{-4mm}
    \caption{Comparison with a 3E6 budget in CEC2013LSGO.}
    \label{CSG, CC-CMAES, ERDG, and LCC at Different FEs}
\end{figure}


\begin{table}[t]
  \centering
  \tiny
  \vspace{+3mm}
  \caption{Comparison of resource consumption.}
  \setlength{\tabcolsep}{2pt}
  \resizebox{0.7\textwidth}{!}{ 
  \vspace{-3mm}
    \begin{tabular}{lccccccc}
    \toprule
    \multicolumn{1}{c}{Algorithm} & LCC-CMAES & CC-CMAES & CSG & ERDG & MDG & FII & CMA-ES \\
    \midrule
    \multicolumn{1}{c}{FEs} & \textbf{0} & \textbf{0} & 4.86E+04 & 1.28E+05 & 4.10E+03 & 4.52E+03 & \textbf{0}\\
    Time Cost (s) & 85.12 & \textbf{83.88} & 275.05 & 173.75 & 332.32 & 264.81 & 450.4\\
    \bottomrule
    \end{tabular} 
    }
    \label{Comparison Of Resource Consumption_1}%
\end{table}%

\subsubsection{Comparison of Resource Consumption} \label{Comparison Of Resource Consumption}
In the analysis above, LCC-CMAES demonstrated its capability to handle more complex problems with a small budget. Besides the additional FEs, the time cost for decomposition and optimization is another bottleneck that constrains the algorithm from being applied to more complex and higher-dimensional problems. Therefore, we conducted a more thorough investigation into the additional FEs and the time cost for decomposition and optimization of each algorithm, aiming for a more detailed presentation of resource consumption.

Tables \ref{Comparison Of Resource Consumption_1} show the additional FEs for decomposition and the averaged optimization time cost (in seconds) for each problem and each run. To avoid the issue of additional FEs becoming excessively high due to certain problems being difficult to decompose, we use the median to reflect the additional FEs. 
A notable advantage of LCC-CMAES is that it does not require additional FEs for decomposition, and due to its simple actor and critic network design, the time expenditure for LCC-CMAES is relatively low. This provides a feasible approach for solving more complex and higher-dimensional real-world problems. Other algorithms under CC framework often suffer from excessive costs due to the difficulty in identifying separable types; even when problems are identifiable, the presence of overlapping issues can result in many subgroups remaining too large, thereby causing substantial time expenditures for the underlying optimizer. Besides, we note that the algorithm without CC usually require more optimization time than CC methods. For instance, CC methods based on CMA-ES are faster than CMA-ES without CC. It is contributed by the decomposed smaller subspace dimensions, which validates the necessary of problem decomposition and CC.

\subsection{The Transferability Study}\label{Transfer}

To more effectively test the transferability of LCC-CMAES, we tested using an entirely new set of problems that it had never encountered before with the same settings. The majority of separable functions in the CEC 2013 LSGO are additively separable, with the Ackley function being the only non-additively separable function among the basic functions. To address these limitations, the BNS \cite{chen2022decomposition}  introduces four non-additively separable base functions, including two multiplicatively separable base functions and two composite separable base functions. Based on these basis functions, BNS designs 12 test problems with varying degrees of separability. Compared to CEC 2013 LSGO, the problems in BNS are closer to the potential complex problems encountered in real-world scenarios. 
All algorithm settings are consistent with those used in the comparison in CEC2013 LSGO.

\begin{table}[t]
  \centering
  \tiny
  \caption{Comparing LCC-CMAES with comparison algorithms on BNS.}
  \resizebox{\textwidth}{!}{ 
    \begin{tabular}{rcccccccccc}
    \toprule
    \multicolumn{1}{c}{problem} & LCC-CMAES & CC-CMAES & CSG & ERDG & MDG & FII & CMA-ES & Sep-CMAES & LM-MA-ES & LM-CMA\\
    \midrule
    \multicolumn{1}{c}{1} & 4.752E-08 & 1.901E-07(+) & 1.741E-06(+) & 3.305E+06(+) & 4.463E-03(+) & 7.022E-04(+) & 4.924E-11(-) &\textbf{0.000E+00}(-) & \textbf{0.000E+00}(-) & \textbf{0.000E+00}(-)\\
      & $\pm$3.125E-08 & $\pm$1.523E-07 & $\pm$9.037E-07 & $\pm$1.905E+06 & $\pm$2.578E-03 & $\pm$8.325E-04 & $\pm$6.043E-11 &$\pm$\textbf{0.000E+00} & $\pm$\textbf{0.000E+00} & $\pm$\textbf{0.000E+00} \\
    \multicolumn{1}{c}{2} & 7.044E+00 & \textbf{6.947E+00}($\approx$) & 1.035E+01(+) & 1.198E+11(+) & 1.185E+01(+) & 1.432E+01(+) & 1.038E+01(+) & 2.375E+02(+)  & 8.306E+03(+) & 6.403E+01(+)\\
      & $\pm$4.794E-01 & $\pm$\textbf{2.413E+00} & $\pm$1.532E+00 & $\pm$5.394E+09 & $\pm$3.865E+00 & $\pm$4.413E+00 & $\pm$4.152E-01 & $\pm$6.350E+01 & $\pm$4.066E+03 & $\pm$1.583+01 \\
    \multicolumn{1}{c}{3} & 7.292E+05 & 5.375E+05($\approx$) & 7.173E+05($\approx$) & 8.092E+06 (+) & 2.569E+06(+) & 7.136E+05($\approx$) & 8.565E+05(+) & \textbf{8.361E+02}(-) & 3.472E+06(+) & 2.881E+06(+) \\
      & $\pm$3.484E+05& $\pm$2.567E+05 & $\pm$3.925E+05 & $\pm$4.024E+04 & $\pm$2.493E+06& $\pm$3.935E+05 & $\pm$1.889E+05 &$\pm$\textbf{4.656E+01} & $\pm$2.455E+05 & $\pm$1.323E+05\\
    \multicolumn{1}{c}{4} & \textbf{1.423E+09} & 4.124E+09(+) & 2.545E+10(+) & 6.812E+11(+) & 1.389E+11($\approx$) & 8.347E+10(+) & 1.852E+10(+) &3.865E+09(+)&7.713E+09(+)&9.296E+09(+)  \\
      & $\pm$\textbf{3.39E+08} & $\pm$1.484E+08 & $\pm$8.115E+08 & $\pm$5.046E+09 & $\pm$7.765E+10 & $\pm$4.438E+09 & $\pm$6.297E+08 & $\pm$2.779E+08 & $\pm$2.510E+08 & $\pm$5.068E+08 \\
    \multicolumn{1}{c}{5} & \textbf{0.000E+00} & 4.402E-08(+) & 4.414E-02(+) & 3.682E+06(+) & 5.003E-02(+) & 2.624E-02(+) & 4.225E-11(+) &\textbf{0.000E+00}($\approx$)  &\textbf{0.000E+00}($\approx$) & \textbf{0.000E+00}($\approx$)\\
      & $\pm$\textbf{0.000E+00} & $\pm$4.876E-08 & $\pm$5.875E-03 & $\pm$2.384E+06 & $\pm$1.665E-03 & $\pm$5.914E-03 & $\pm$6.183E-11 & $\pm$\textbf{0.000E+00} & $\pm$\textbf{0.000E+00} & $\pm$\textbf{0.000E+00} \\
    \multicolumn{1}{c}{6} & 2.105E+01 & 4.084E+01(+) & 2.278E+03(+) & 1.189E+11 (+) & 5.694E+04(+) & 2.366E+04(+) & \textbf{8.482E+00}(-) & 2.763E+02(+) & 4.035E+01(+) & 3.913E+02(+)\\
      & $\pm$3.663E+00 & $\pm$9.414E+00 & $\pm$1.385E+03 & $\pm$4.286E+10 & $\pm$9.847E+03 & $\pm$7.106E+03 & $\pm$\textbf{1.185E+00} & $\pm$2.884E+01 & $\pm$5.894E+00  & $\pm$3.653E+01 \\
    \multicolumn{1}{c}{7} & 3.841E+06 & 3.846E+06($\approx$) & 5.554E+06(+) & 8.095E+06(+) & 6.334E+06(+) & 6.376E+06(+) & \textbf{9.947E+05}(-) & 2.673E+06(-) & 2.829E+06(-) & 3.845E+06($\approx$)\\
      & $\pm$3.052E+05 & $\pm$9.843E+04 & $\pm$5.645E+04 & $\pm$5.256E+03 & $\pm$4.891E+04 & $\pm$4.795E+04 & $\pm$\textbf{1.596E+05} &$\pm$2.435E+05 & $\pm$1.932E+05 & $\pm$6.532E+05\\
    \multicolumn{1}{c}{8} & 1.982E+10 & 2.423E+10(+) & 1.844E+11(+) & 6.705E+11(+) & 1.085E+11(+) & 1.137E+11(+) & 1.932E+10($\approx$) & 2.674E+10(+) & 8.473E+09(-) & \textbf{4.77E+09}(-)\\
      & $\pm$1.124E+09 & $\pm$1.523E+09 & $\pm$1.804E+10 & $\pm$5.125E+10 & $\pm$5.472E+09 & $\pm$5.171E+09 & $\pm$4.945E+08 & $\pm$1.142E+09 &$\pm$5.432E+08 & $\pm$\textbf{5.223E+08} \\
    \multicolumn{1}{c}{9} & 4.883E-08 & 2.094E-07(+) & 7.395E-04(+) & 1.443E+07(+) & 5.534E-03(+) & 3.975E-02(+) & 4.452E-08($\approx$) & \textbf{0.000E+00}(-) & \textbf{0.000E+00}(-)&\textbf{0.000E+00}(-)\\
      & $\pm$1.705E-08 & $\pm$1.467E-07 & $\pm$2.178E-04 & $\pm$8.453E+04 & $\pm$1.115E-03 & $\pm$4.973E-03 & $\pm$1.775E-08 & $\pm$\textbf{0.000E+00}& $\pm$\textbf{0.000E+00} &$\pm$\textbf{0.000E+00}\\
    \multicolumn{1}{c}{10} & 3.074E+02 & 4.213E+04(+) & 3.725E+04(+) & 7.054E+10(+) & 1.635E+06(+) & 1.713E+05(+) & \textbf{1.062E+01}(-) & 2.943E+02($\approx$) & 5.309E+01(-) & 2.154E+03(+)  \\
      & $\pm$2.901E+02 & $\pm$3.372E+04 & $\pm$9.991E+03 & $\pm$3.705E+09 & $\pm$6.398E+05 & $\pm$7.537E+04 & $\pm$\textbf{8.083E-01} & $\pm$9.455E+01 & $\pm$1.618E+01 & $\pm$1.401E+02 \\
    \multicolumn{1}{c}{11} & 3.762E+06 & 4.062E+06(+) & \textbf{3.506E+06}($\approx$) & 8.057E+06(+) & 7.228E+06(+) & 7.196E+06(+) & 4.574E+06(+) & 3.475E+06($\approx$) & 6.021E+06(+) & 3.777E+06($\approx$)\\
      & $\pm$2.562E+05 & $\pm$1.235E+05 & $\pm$\textbf{7.774E+04} & $\pm$3.916E+04 &$\pm$4.735E+04 & $\pm$4.051E+04 & $\pm$2.962E+05 & $\pm$5.456E+05 & $\pm$6.778E+04 & $\pm$6.542E+05   \\
    \multicolumn{1}{c}{12} & 2.821E+10 & 4.289E+10(+) & 5.365E+10(+) & 6.782E+11(+) & 1.952E+10(-) & 2.075E+10(-) & 1.911E+10(-) & 3.913E+10(+) &2.782E+10(-) & \textbf{4.612E+09}(-) \\
      & $\pm$1.632E+09 & $\pm$2.254E+09 & $\pm$1.965E+09 & $\pm$3.076E+10 & $\pm$8.082E+08 & $\pm$1.494E+09 & $\pm$7.118E+08 & $\pm$2.371E+09 & $\pm$2.407E+09 & $\pm$\textbf{4.578+E08}\\
    \midrule
      & NA & 9/3/0 & 10/2/0 & 12/0/0 & 10/1/1 & 10/1/1 & 5/2/5 & 5/3/4 & 5/2/5 &5/3/4\\
    \bottomrule
    \end{tabular}%
    }
  \label{BNS}%
\end{table}%

Table \ref{BNS} presents the comparative results, indicating that LCC-CMAES has certain advantages. On one hand, some algorithms consume excessively high additional FEs for decomposition on BNS problems, which prevents them from focusing resources on the optimization process, leading to poor performance (e.g., ERDG). On the other hand, some algorithms fail to correctly identify variable interactions and group them on more complex problems, resulting in poor performance (e.g., MDG, with almost zero correct grouping rate). These results also reveal the transferability of LCC-CMAES and its potential in tackling more complex real-world problems: trained on simpler benchmarks, it can transfer decomposition knowledge to more complex real scenarios, discovering grouping structures through learning rather than relying on expert-level knowledge. However, LSGO variant algorithms and CMA-ES don't encounter issues with inaccurate decomposition and additional FEs, and it can fully leverage the relationships between variables. This underscores the challenge that CC faces in thousands to low dimensions compared to global optimizers.

\subsection{Comparison on Neuroevolution tasks}\label{sec:ne}

In this section, we adopt four Neuroevolution~\cite{neuralevolution} tasks as a showcase on real-world applications, in which optimization algorithms are used to evolve a population of neural networks according to their performance on a specific machine learning task such as robotic control~\cite{neuroevolve-survey}. Concretely, we consider the real-parameter optimization of 2-layer MLPs for 4 Mujoco~\cite{mujoco} robot control tasks: \textit{InvertedDoublePendulum-v4}, \textit{HalfCheetah-v4}, \textit{Pusher-v4} and \textit{Ant-v4}. We set the hidden dimensions of the MLPs to 64 with Tanh activation function, while the input and output dimensions match the control protocols of the Mujoco tasks, leading to 833, 1542, 1991 and 2312 dimensions for the four MLPs of the four tasks, respectively. Because the evaluations of the networks are time consuming, we set the maximum function evaluations (FEs) of all tasks to 1,000. For baselines, we zero-shot the pre-trained LCC-CMAES agent in Section 5.2 to the Neuroevolution problems. The CC-based baselines except CC-CMAES fail to decompose the problem dimensions within 1,000 FEs so we do not include them in the comparison. The hyper-parameter settings of LCC-CMAES and included baselines are consistency with Section 5.1.2 except the total number of generations ($TG$) and the offspring size ($\lambda$) of LCC-CMAES which are both set to 5.
Since the targets in the Mujoco tasks are to maximize the accumulated rewards gained by the networks, in Table~\ref{tab:ne} we present the negative accumulated rewards obtained by the networks optimized by LCC-CMAES and baselines to keep the minimization optimization manner. 

The results show that even on realistic optimization problems with larger problem dimensions and more complex variable relationship, our LCC-CMAES still retains its advantages over CC-CMAES and global optimization baselines, validating its effectiveness.

\begin{table}[t]
\centering
\caption{Comparison results on Neuroevolution tasks.}
\label{tab:ne}
\resizebox{\columnwidth}{!}{%
\begin{tabular}{c|cccccccc}
\hline
 &
  LCC-CMAES &
  CC-CMAES &
  CMA-ES &
  Sep-CMAES &
  LM-CMA &
  LM-MA-ES &
  MetaES &
  L-BFGS \\ \hline
\begin{tabular}[c]{@{}c@{}}InvertedDoublePendulum-v4\\(833D)\end{tabular} &
  \textbf{\begin{tabular}[c]{@{}c@{}}-5.111E+03\\ $\pm$2.914E+02\end{tabular}} &
  \begin{tabular}[c]{@{}c@{}}-4.854E+03 (+)\\ $\pm$2.354E+02\end{tabular} &
  \begin{tabular}[c]{@{}c@{}}-4.714E+03 (+)\\ $\pm$2.354E+02\end{tabular} &
  \begin{tabular}[c]{@{}c@{}}-4.988E+03 (+)\\ $\pm$2.644E+02\end{tabular} &
  \begin{tabular}[c]{@{}c@{}}-4.971E+03 (+)\\ $\pm$2.541E+02\end{tabular} &
  \begin{tabular}[c]{@{}c@{}}-4.951E+03 (+)\\ $\pm$2.455E+02\end{tabular} &
  \begin{tabular}[c]{@{}c@{}}-4.596E+03 (+)\\ $\pm$2.944E+02\end{tabular} &
  \begin{tabular}[c]{@{}c@{}}-4.322E+03 (+)\\ $\pm$2.831E+02\end{tabular} \\ \hline
\begin{tabular}[c]{@{}c@{}}HalfCheetah-v4\\(1542D)\end{tabular} &
  \textbf{\begin{tabular}[c]{@{}c@{}}-2.451E+02\\ $\pm$2.514E+02\end{tabular}} &
  \begin{tabular}[c]{@{}c@{}}-2.017E+02 (+)\\ $\pm$2.119E+02\end{tabular} &
  \begin{tabular}[c]{@{}c@{}}-1.914E+02 (+)\\ $\pm$1.645E+02\end{tabular} &
  \begin{tabular}[c]{@{}c@{}}-1.849E+02 (+)\\ $\pm$2.002E+02\end{tabular} &
  \begin{tabular}[c]{@{}c@{}}-1.897E+02 (+)\\ $\pm$2.129E+02\end{tabular} &
  \begin{tabular}[c]{@{}c@{}}-1.744E+02 (+)\\ $\pm$1.988E+02\end{tabular} &
  \begin{tabular}[c]{@{}c@{}}-5.554E+01 (+)\\ $\pm$9.624E+01\end{tabular} &
  \begin{tabular}[c]{@{}c@{}}-4.997E+01 (+)\\ $\pm$9.487E+01\end{tabular} \\ \hline
\begin{tabular}[c]{@{}c@{}}Pusher-v4\\(1991D)\end{tabular} &
  \begin{tabular}[c]{@{}c@{}}3.354E+02\\ $\pm$2.984E+01\end{tabular} &
  \begin{tabular}[c]{@{}c@{}}3.543E+02 (+)\\ $\pm$3.791E+01\end{tabular} &
  \begin{tabular}[c]{@{}c@{}}3.497E+02 (+)\\ $\pm$4.016E+01\end{tabular} &
  \begin{tabular}[c]{@{}c@{}}3.481E+02 (+)\\ $\pm$3.594E+01\end{tabular} &
  \textbf{\begin{tabular}[c]{@{}c@{}}3.344E+02 (-)\\ $\pm$2.326E+01\end{tabular}} &
  \begin{tabular}[c]{@{}c@{}}3.411E+02 (+)\\ $\pm$2.746E+01\end{tabular} &
  \begin{tabular}[c]{@{}c@{}}3.894E+02 (+)\\ $\pm$4.687E+01\end{tabular} &
  \begin{tabular}[c]{@{}c@{}}3.909E+02 (+)\\ $\pm$6.314E+01\end{tabular} \\ \hline
\begin{tabular}[c]{@{}c@{}}Ant-v4\\(2312D)\end{tabular} &
  \textbf{\begin{tabular}[c]{@{}c@{}}-1.083E+03\\ $\pm$6.476E+01\end{tabular}} &
  \begin{tabular}[c]{@{}c@{}}-1.080E+03 ($\approx$)\\ $\pm$6.524E+01\end{tabular} &
  \begin{tabular}[c]{@{}c@{}}-1.073E+03 (+)\\ $\pm$5.146E+01\end{tabular} &
  \begin{tabular}[c]{@{}c@{}}-1.066E+03 (+)\\ $\pm$5.687E+01\end{tabular} &
  \begin{tabular}[c]{@{}c@{}}-1.085E+03 ($\approx$)\\ $\pm$6.971E+01\end{tabular} &
  \begin{tabular}[c]{@{}c@{}}-1.079E+03 ($\approx$)\\ $\pm$5.377E+01\end{tabular} &
  \begin{tabular}[c]{@{}c@{}}-1.015E+03 (+)\\ $\pm$3.345E+01\end{tabular} &
  \begin{tabular}[c]{@{}c@{}}-1.004E+03 (+)\\ $\pm$2.154E+01\end{tabular} \\ \hline
    & NA & 3/1/0 & 4/0/0 & 4/0/0 & 2/1/1 & 3/1/0 & 4/0/0 & 4/0/0 \\
    \hline
\end{tabular}%
}
\end{table}

\subsection{Ablation Study}\label{Ablation Studie}

\subsubsection{State features Study} \label{State features Study}

To verify the necessity of the framework components, we conducted ablation studies on the state features. Specifically, we separately removed the embeddings and concatenations of the $s_{\text{AH}}$ (denoted as W/O AH), $s_{\text{GO}}$ (W/O GO), and $s_{\text{SD}}$ (W/O SD). We then tested these modifications on the CEC 2013 LSGO under the same settings otherwise unchanged.


The results are shown in Figure \ref{The ablation study performance} (A). 
For each problem, we use the average performance of all runs for each algorithm and conduct the min-max normalization over all algorithms to restrict their performance into $[0, 1]$ and eliminate the cost scale gaps between different problems. The $1 -$ mean performance over all problems of each algorithms and their error bars are presented, where the higher is better.
It is evident that the performance significantly deteriorates when these features are removed, highlighting their crucial roles to provide sufficient information being available to the RL agent.

\subsubsection{Reward Study}

The design of the reward mechanism needs to employ a ratio-based approach within the range of -1 to 1 to address that evaluation values can vary significantly across different problems and excessive impacts on network training. Here are two other reward designs that meet these requirements:

1) Global best fitness descent ratio: The global best fitness is calculated by subtracting it from the initial generation global best fitness and normalizing the decline by the global best fitness of the initial generation. reward1 := $r_{t}=\frac{f_{0}^{*}-f _{t}^{*}}{f_{0}^{*}}$

2) Relative global best fitness descent ratio: The global best fitness decline normalized by the global best fitness in previous generation. reward2 := $r_{t}=\frac{f _{t-1}^{*}-f _{t}^{*}}{f _{t-1}^{*}}$

\begin{figure}[t]
    \centering
    \includegraphics[width=1\textwidth]{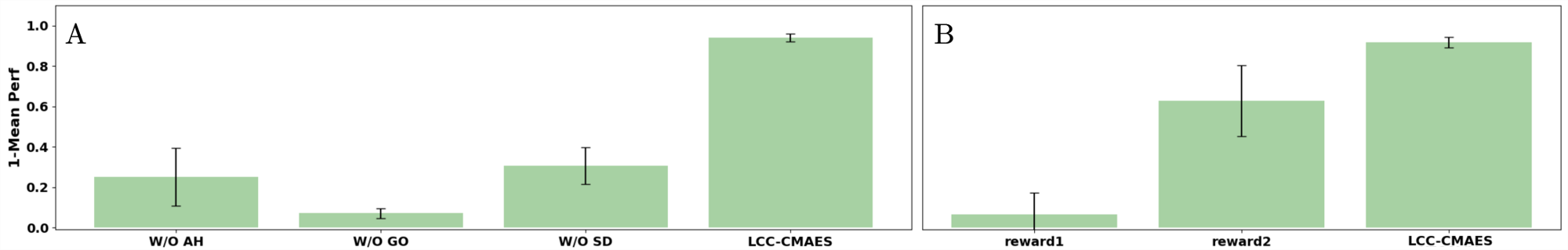}
    \caption{The ablation study on state features and reward designs.}
    \label{The ablation study performance}
\end{figure}

Figure \ref{The ablation study performance} (B) also presents the results under different reward schemes, with the same normalization as introduced in Section \ref{State features Study}. It can be observed that the reward1 and the reward2 are significantly less effective than the approach we have adopted. This ineffectiveness is due to the fact that using a scheme that subtracts the initial generation's fitness can lead to many subsequent fitness values significantly lower than the initial generation, making the numerator approximately equal to the initial value, causing the formula to approach 1. Additionally, normalizing the global best fitness from the previous generation causes the standard for reward normalization to change with each generation. This inconsistency makes it challenging for the RL agent to select appropriate actions.

\section{Conclusion and Future work}\label{Conclusion and Futrue work}

We have proposed LCC, a pioneering learning-based cooperative coevolution framework that dynamically schedules decomposition strategies during optimization processes. With CMA-ES as the underlying optimizer, we instantiate LCC, naming it the LCC-CMAES algorithm. Unlike previous algorithms under the CC framework, LCC-CMAES does not use expert-designed knowledge for decomposition but instead utilizes statistical features for DRL to select most-expected decomposition strategies. More importantly, LCC-CMAES does not require the additional FEs for decomposition, allowing it to focus resources on optimization. When tested against several other advanced algorithms on two benchmarks, CEC 2013 LSGO and BNS, the comparative results demonstrated that LCC-CMAES holds a distinct advantage, especially on complex real-world problems that it had not previously encountered. This underscores LCC's robustness, adaptability and transferability, making it a promising approach for tackling complex optimization challenges in various settings.

Looking ahead to future work, we hope to: (1) investigate the inclusion of more complex or higher-dimensional features that may capture deeper insights into the problem’s structure; (2) design more rational and effective decomposition actions. These goals aim to refine LCC's effectiveness and applicability, ensuring it can be a versatile tool in the LSGO, capable of addressing a broader range of complex challenges.

\begin{acks}
This work was supported in part by the National Natural Science Foundation of China No. 62276100, in part by the Guangdong Provincial Natural Science Foundation for Outstanding Youth Team Project No. 2024B1515040010, in part by the Guangdong Natural Science Funds for Distinguished Young Scholars No. 2022B1515020049, and in part by the TCL Young Scholars Program.
\end{acks}

\bibliographystyle{ACM-Reference-Format}
\bibliography{sample-base}

\end{document}